# Overcoming Overfitting and Large Weight Update Problem in Linear Rectifiers: Thresholded Exponential Rectified Linear Units

Vijay Pandey[1]

[1]Department of Computer Science Engineering, IIT Kharagpur, India

E-mail ID: vijayiitkgp13@gmail.com

## ABSTRACT

In past few years, linear rectified unit activation functions have shown its significance in the neural networks, surpassing the performance of sigmoid activations. RELU (Nair & Hinton, 2010), ELU (Clevert et al., 2015), PRELU (He et al., 2015), LRELU (Maas et al., 2013), SRELU (Jin et al., 2016), *ThresholdedRELU,* all these linear rectified activation functions have its own significance over others in some aspect. Most of the time these activation functions suffer from bias shift problem due to non-zero output mean, and high weight update problem in deep complex networks due to unit gradient, which results in slower training, and high variance in model prediction respectively. In this paper, we propose, "Thresholded exponential rectified linear unit" (TERELU) activation function that works better in alleviating in overfitting: large weight update problem. Along with alleviating overfitting problem, this method also gives good amount of non-linearity as compared to other linear rectifiers. We will show better performance on the various datasets using neural networks, considering TERELU activation method compared to other activations.

## 1. INTRODUCTION

Activation function is the key component in the neural networks. Activation function is primarily used to provide non-linearity in the representation. As the rapid evolvement of the neural networks in past few years happened, number of researches have been carried out in this field. In earlier days, sigmoid activation functions were very popular. These functions are symmetrical; provide good non-linearity in the representation. They have two extreme points for neuron saturation (deactivation). *Logistic sigmoid* was replaced by *tanh sigmoid* because *tanh* function has output mean zero (Y. A. LeCun et al., 2012). On the other hand *logistic sigmoid* does not has output mean as zero. Activation functions having zero mean output, show faster learning (Clevert et al., 2015; Schraudolph, 1998). It has been shown that zero mean is very helpful in many ways. The gap between natural gradient and normal gradient is reduced, which helps in learning network fast (Clevert et al., 2015). These all benefits come with some problem. The main problem is the vanishing gradient problem (Hochreiter et al., 2001). Sigmoid activations saturates on their extreme asymptotes, which leads to very small gradients after few training iterations in network. Due to vanishing gradient problem, lower layer weights stop learning and, as result model does not learn well (Nwankpa et al., 2018).

*Linear Rectifiers* came into popularity to overcome on the vanishing gradient problem in *sigmoid activations* (Glorot et al., 2011; Hochreiter, 1998). Linear rectifiers have unit gradient for positive value of input so, gradient does not vanish for deep networks. Rectifiers suffer from the bias shift problem due to non-zero mean output (Clevert et al., 2015). Along with it, rectifiers in general does not provide the good non-linearity in the activation output as sigmoid activations provide. At the origin, change in slope is sudden, not smooth for linear rectifiers.

In this paper, new activation function *TERELU* is proposed, which is, based on the linear rectifiers and sigmoid activation. *TERELU* takes advantages of both sigmoid activation function as well as linear rectifiers. *tanh* Sigmoid activations were very popular in earlier days due to its property of extreme asymptote at some range (-1, 1), which produces zero-mean activation. The one of the main advantage of zero mean is, bias shift value is small and the output is centred at zero mean, which works as normalizing the data, having zero mean and unit variance. The main problem with sigmoid activations is, they become saturated on their extreme asymptotes, which leads to very small gradients after few iteration in network training. Due to small gradients, network suffer from vanishing gradient problem, as result lower layer weights stops learning, and model does not learn well. Rectifiers came into popularity to overcome the steepest gradient problem at its extremes, which helps to learn very deep neural networks. In *RELU*, for positive input value gradient is one and for rest, gradient is zero (Nair & Hinton, 2010). This approach helps to learn the deep model due to positive high gradient even for high input value and thus alleviate the vanishing gradient problem. However, at the same time it suffers from dead neuron problem (Lu et al., 2019). The output mean of *RELU* is always positive because it outputs only positive activations. *LRELU*, *ELU*, *PRELU* help to alleviate the problem of dead neurons. Nevertheless, for all the positive input values, any saturation point is not defined for these activation functions. For positive input values, these activations are purely linear, whether it is R*ELU*, *ELU*, *LRELU*, S*ELU* or *PRELU*. Due to absence of any saturation point defined for all positive values, variants of the R*ELU* also suffers from non-zero mean problem. A method *TERELU* is proposed to solve the above-mentioned problem with linear rectifiers. An exponential function is added for positive inputs greater than user-defined threshold as well on the *ELU* activation.

There are various benefits of giving extra exponential function in *TERELU* as compared to *ELU* activation, 1) It provides saturation point for positive values of inputs as well, which helps in keeping the mean output towards zero. 2) It helps in keeping the bias shift value low for each layer, which help in faster network learning. 3) It helps in providing good non-linearity as compared to the only linear values. 4) Helps in alleviating the large weight updates for the deep network due to accumulation of large error gradients of unit value.

## 2. RELATED WORK

1. Rectified Linear Unit (RELU) (Nair & Hinton, 2010)

$$f(x) = \begin{cases} 0, & x \leq 0 \\ x, & x > 0 \end{cases} \quad (1)$$

2. Leaky RELU (Maas et al., 2013)

$$f(x) = \begin{cases} \alpha x, & x \leq 0 \\ x, & x > 0 \end{cases} \quad (2)$$

It is the special version of a RELU that allows a small gradient when the unit is not active. $\alpha \geq 0$, which is negative slope coefficient.

3. Exponential linear units (ELU) (Clevert et al., 2015)

$$f(x) = \begin{cases} \alpha(e^x - 1), & x \leq 0 \\ x, & x > 0 \end{cases} \quad (3)$$

Here $\alpha \geq 0$ is the scale for the negative factor. ELU activation is depicted in figure 1.

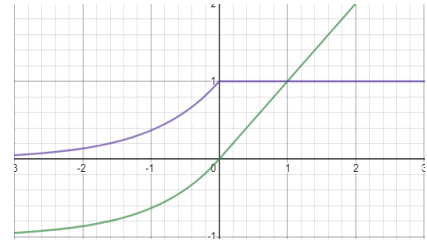

Figure 1: ELU activation in green curve, its derivative in purple curve.

4. SRELU (Jin et al., 2016)

$$f(x) = \begin{cases} t_i^r + a_i^r(x_i - t_i^r), & x_i \geq t_i^r \\ x_i, & t_i^r \geq x_i \geq t_i^l \\ t_i^l + a_i^l(x_i - t_i^l), & x_i \leq t_i^l \end{cases} \quad (4)$$

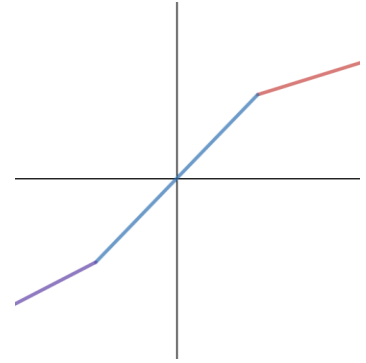

Figure 2: SRELU activation function.

Where $\{t_i^r, a_i^r, t_i^l, a_i^l\}$ are four learnable parameters used to model an individual *SRELU* activation unit. The subscript $i$ indicates that we allow *SRELU* to vary in different channels.

5. Adaptive Piecewise Linear Units (APL) (Agostinelli et al., 2014)

The method formulates the activation function $h_i(x)$ of an APL unit $i$ as a sum of hinge-shaped functions,

$$h_i(x) = \max(0, x) + \sum_{s=1}^{S} a_i^s \max(0, -x + b_i^s) \quad (5)$$

This is a piecewise linear function. S is the number of hinges, which is a hyperparameter, while other variables are learned using standard gradient descent during training. $a_i^s$ Controls the slope of the linear segments and $b_i^s$ determine the location of the hinges.

6. MAXOUT (Goodfellow et al., 2013)

Maxout is a feed-forward architecture that uses a new type of activation function: the maxout unit. Given an input $x \in R^d$, $x$ may be v, or it may be hidden layer's state. Maxout hidden layer implements the function,

$$h_i(x) = \max_{j \in [1,k]} z_{ij} \quad (6)$$

Where $z_{ij} = x^T W_{...ij} + b_{ij}$, $W \in R^{d \times m \times k}$ and $b \in R^{m \times k}$ are learned parameters.

7. ParametricSoftplus (McFarland et al., 2013)

Parametric softplus activation function is given by,

$$\alpha * \log(1 + e^{\beta x}) \quad (7)$$

8. Tanh activation

It is one of the popular activation in earlier days, formulation of tanh activation is,

$$f(x) = \frac{e^x - e^{-x}}{e^x + e^{-x}} \qquad (8)$$

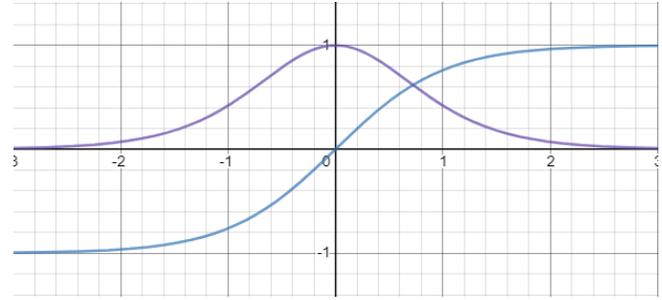

Figure 3: Tanh activation in blue curve, its derivative in purple curve.

## 3. OVERFITTING AND LARGE WEIGHT UPDATE PROBLEM IN LINEAR RECTIFIERS FOR DEEP NETWORKS

Experiment conducted in this paper shows that linear rectifier suffer from overfitting problem, in deep networks. It is shown in fig. 1. One main reason for this is no provision of contractive gradients for positive inputs. Suppose in a given scenario, all the input values are positive. If gradient value is of the same sign then the weight change will move in a single direction, either negative or positive, which hampers the learning. If there are some weights, which moves in positive direction having more weight, and the input will be positive as well, and given the unit gradient, then it will keep moving in the positive side without any saturation point for the positive input, this will keep going further as learning progress, without reaching any saturation point for the positive input. This will cause the weight updation, for those neurons, which has the positive activation. This cause the higher weight updates, which makes the learning unstable. This results in overfitting on the given data as training progresses.

There is no provision of weight decay, in the linear activations, as the gradient does not decrease for input $x > 0$ it is unit gradient, which results in over trained weights. Value of few weights keep changing with high updation value to fit the data well. For the higher value of $x$, gradient value will be high, and as result, weight update will be high due to this. It can be in any direction whether weigh will be increased by high value or get decrease by high value. This high updation will happen even in those cases, where model is already trained. So further high updation will result in overfitting and unstable condition of the model. To reduce this, in the proposed method we are giving a saturation point for all those inputs having values $x>0$. Here $\beta$ is working as a regularising parameter, which helps in decaying the weights, for higher value of $x$.

## 4. THRESHOLDED EXPONENTIAL RECTIFIED LINEAR UNIT (TERELU)

Thresholded exponential RELU is given as,

$$f(x) = \begin{cases} \alpha(e^x - 1), & x \leq 0 \\ x, & \mu > x > 0 \\ \beta(\mu - (e^{(-x+\mu)} - 1)), & x \geq \mu \end{cases} \quad (9)$$

Derivative of the above activation function is given as,

$$f'(x) = \begin{cases} f(x) + \alpha, & x \leq 0 \\ 1, & \mu > x > 0 \\ -f(x) + \beta\mu + \beta, & x \geq \mu \end{cases} \quad (10)$$

Where $\alpha > 0, \mu > 0$ are hyperparameters. Parameter $\beta$ is trainable parameter. Default value for $\alpha, \beta$ and $\mu$ is one.

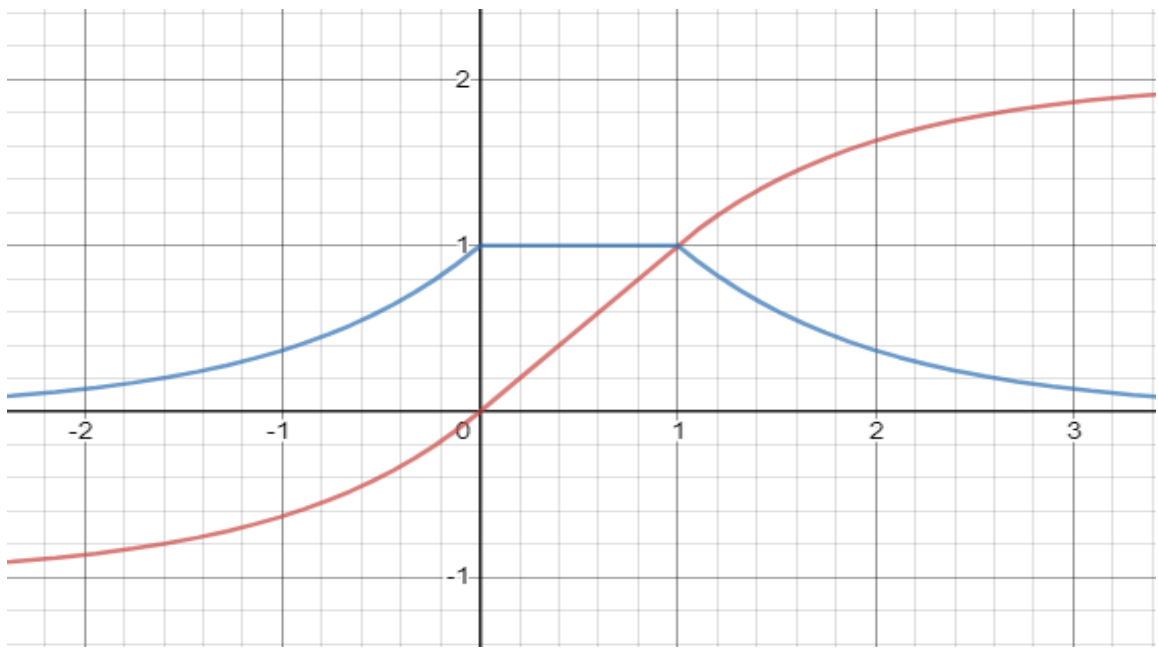

Figure 4: TERELU activation. Red curve represents activation. Blue curve represent its derivation.

Here μ works to adjust the value of $x$, below which activation will work as linear rectifiers, and β is working as a regularizer (Rifai et al., 2011) and training accelerator as well, which increases and decreases the gradient value as per the training. Decreasing gradient value work as the weight decay (Rifai et al., 2011). Increasing gradient value help in learning, and unlike the sigmoid activations, helps network to learn continuously, even for higher positive inputs, without being saturated, in subsequent iterations. *TERELU* also helps in normalizing the data well in contrast with other variant of linear rectifiers. Normalizing the output activation has a very significant role in neural network training (Ioffe & Szegedy, 2015; Santurkar et al., 2018). *TERELU* also helps in providing the sparsity of data and in creating sparse representations, by saturating the neuron of higher activations, which helps in giving better performance (Glorot et al., 2011; Maas et al., 2013).

The saturation decreases the variation of the units if deactivated, so deactivation value does not have high significance as compared with activation value and thus is less relevant. Such an activation function can code the degree of presence of particular phenomena in the input, but does not quantitatively model the degree of their absence (Clevert et al., 2015). Therefore, such an activation function is more robust to noise.

## 5. ADDING EXPONENTIAL FUNCTION FOR HIGHER POSITIVE INPUTS INCREASES GENERALISATION POWER IN LINEAR RECTIFIERS

For deep neural networks, linear rectifiers show *exploding gradients* problem at some extent, where large error gradients accumulate, which results in very large updates to neural network model weights during training (Pascanu et al., 2013). This occurs through exponential growth by repeatedly multiplying gradients through the network layers that have values not less than 1.0. Side effect of this behaviour is model becomes unstable and unable to learn from your training data. In *TERELU* due to added exponential function for positive value inputs, it works as contracting gradients for higher input values, thus working as regularizer to achieve weight decay (Rifai et al., 2011). It helps in overcoming the problem of accumulation of non-contractive gradients, which propagates through large number of hidden layers (Bengio et al., 1994).

## 6. COMPARISON WITH OTHER ACTIVATION FUNCTIONS

In linear rectifier based activation, activation layer always give positive unit gradient during back propagation for positive output value except *SRELU*. There is no flexibility for training process to change the gradient as per the need while training, to give better result. In *SRELU* there is some provision provided for the same, however the defined activation is not very smooth and do not ensure a noise-robust deactivation state. We can also see the bias shift in *TERELU* is low as compared to the *ELU* due to using exponential function at both ends i.e. for negative (below 0) as well as positive (above µ) extremes. This helps in reducing the gap between natural gradient (Amari, 1998) and normal gradient, as the activation output approaches to the zero mean, as result non-diagonal entries in the fisher matrix is becoming low. It helps in learning the model fast (Clevert et al., 2015; Raiko et al., 2012; Schraudolph, 1998).

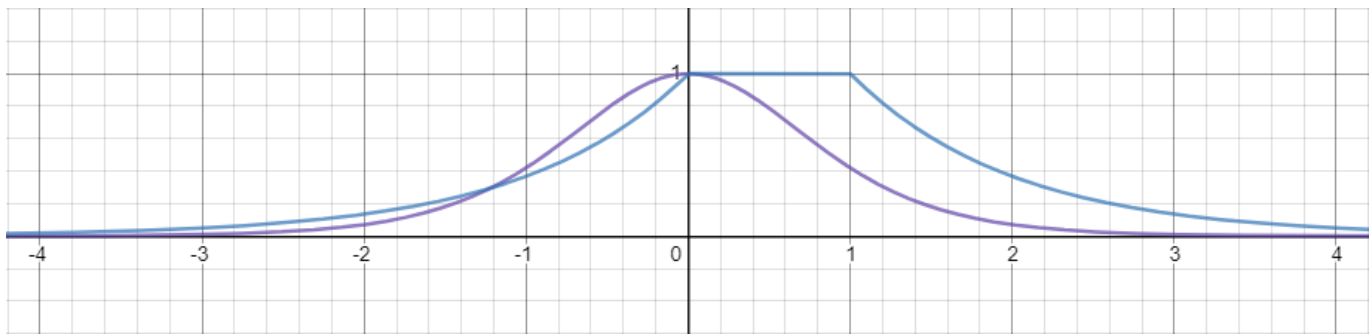

*Figure 5: Tanh function derivative curve in purple colour. TERELU derivative curve in blue curve.*

*TERELU* provide some benefit of sigmoid-based activations by providing saturation point at both negative and positive end and providing good non-linearity, as well as the linear rectifier activations by providing linear activation in between the both extremes, with some hyperparameters and parameter to adjust the weightage of the both type of activations.

Unlike the sigmoidal activation where gradient is almost everywhere contractive, and in *ELU* where gradient is only contractive for negative values of inputs, in TERELU, gradient is contractive only at the extreme i.e. for negative inputs and for positive inputs above some positive threshold $\mu$ and gradient is of unit value in between. Unit gradient helps to flow the relevant information properly.

From the derivation graph of *tanh* and *TERELU*, it can be seen that, in *TERELU* saturation of the positive input is delayed by the threshold parameter $\mu$, as compared to *tanh*. Gradient of *TERELU* starts contracting for positive input after $\mu$ unlike *tanh,* in which gradient is contractive almost everywhere. As a result, *TERELU* does not suffer from vanishing gradient problem as sigmoid activation does for sufficient high value of $\mu$, and lead to robust optimization. *TERELU* does not provide hard-zero sparsity like R*ELU* thus overcoming the problem of dying neuron in R*ELU* (Lu et al., 2019). During training, it has been observed that *TERELU* helps model to converge faster compared to some of the discussed activation functions due to the available saturation point in both directions.

## 7. EXPERIMENT

Experiments on MNIST DATA (Y. LeCun et al., 1998) has been carried out to show the performance of the TERELU over other activation functions. For MNIST data, fully connected neural networks (FCNN) with batch normalization (BN) has been used. Main objective of experiment is to show that TERELU does not over fit on the deep network where linear rectifier based activations does.

### 7.1 MNIST DATA SET

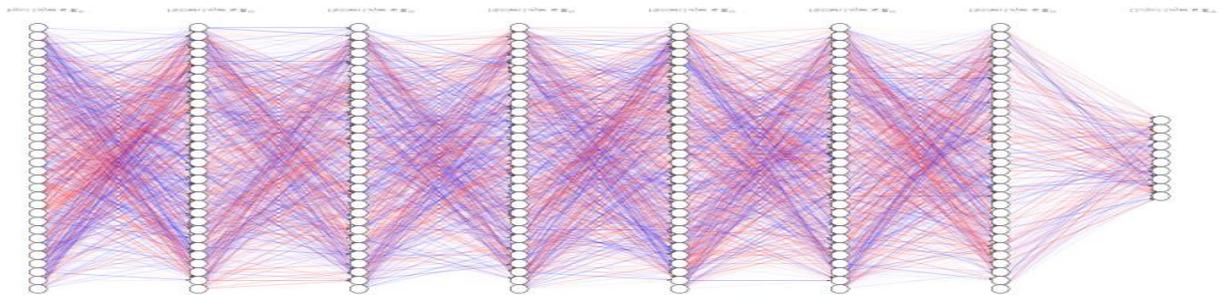

*Figure 6: Fully connected neural network*

54 hidden layers FCNN has been used, along input layer and output layer having 10 units. 8 layer network and 20 layer FCNN has also used for performance measure. See figure 6 for FCNN architecture, figure 6 is for illustration purpose of FCNN architecture. In each above FCNN architecture, number of units are 64 in all hidden layers. Before applying activation function, batch normalization has been used with mean value 0 and unit variance. TERELU performance is better with batch normalization as compared to without BN.

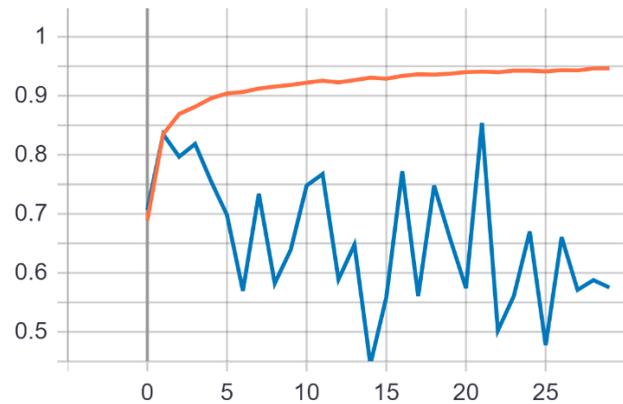
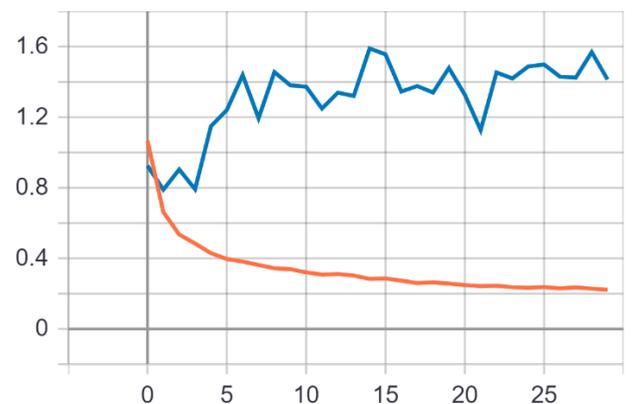

Figure 7: ELU accuracy curve. Orange curve shows training accuracy.

Figure 8: ELU accuracy curve. Orange curve shows training loss.

Experiment has shown that, for 8 layers and 20 layers, both activations ELU and TERELU perform almost very similar, but for the 56 layer FCNN, one can clearly see the huge difference in fig. It can be seen in accuracy curve in figure 7, for ELU activation model starts overfitting and validation error keeps fluctuating with very bad performance. On, the other hand TERELU does not over fit at any point in the curve, figure 9. It converges smoothly with giving training accuracy equal to the validation accuracy. Same trend can be seen in loss curve as well, in given figure 8, validation loss for ELU activation does not decrease after few iteration, while for TERELU validation loss, figure 10, keep decreasing throughout the each iteration.

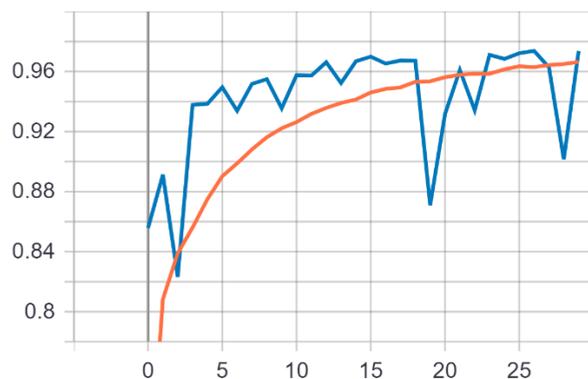
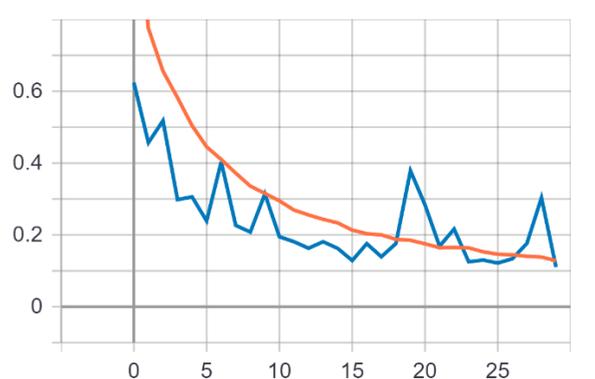

Figure 9: TERELU accuracy curve.

Figure 10: TERELU loss curve.

As from the experiment, it can be seen, for *ELU*: model starts overfitting for deep networks. Flexibility in $\mu$ and $\beta$ is very helpful in learning. Value of $\beta$ is trainable and are adjusted as per the need. Where higher update is needed, $\beta$ value is increased and if lower value is needed, its value is decreased as per the training. Changing value of $\beta$ helps in keep model learning in every iteration, alleviate it from early saturation and stagnation in learning as in sigmoid layers

## 8. CONCLUSION

From the above discussion, it can be seen that due to linear function for positive values of neuron, network suffers from high weight updates, making the training unstable and leading to poor generalization. *TERELU* overcomes form these difficulty by making the gradient value steep as the neuron attains high value. This approach shows the improvement in performance in respect to generalization and makes network training more stable. It can also be seen that *TERELU* can be used for shallow as well as for very deep network to make training more stable and to ensure better normalization of neuron output and generalization over unseen data.